\documentclass[10pt,twocolumn,letterpaper]{article}

\usepackage{cvpr}  
\usepackage{graphicx}
\usepackage{amsmath}
\usepackage{amssymb}
\usepackage{booktabs}
\usepackage{tabu}      
\usepackage{epsfig}
\newcommand\tab[1][1cm]{\hspace*{#1}}

\usepackage[pagebackref,breaklinks,colorlinks]{hyperref}

\usepackage[capitalize]{cleveref}
\crefname{section}{Sec.}{Secs.}
\Crefname{section}{Section}{Sections}
\Crefname{table}{Table}{Tables}
\crefname{table}{Tab.}{Tabs.}

\def\cvprPaperID{17} 
\def\confName{CVPR}
\def\confYear{2022}

\begin{document}

\title{Facial De-occlusion Network for Virtual Telepresence Systems}

\author{
Surabhi Gupta \tab \tab Ashwath Shetty \tab \tab Avinash Sharma\\
\\
{International Institute of Information Technology Hyderabad}\\
{\tt\small \textit {\{surabhi.gupta, ashwath.shetty\}@research.iiit.ac.in}}\\
{\tt\small \textit {asharma@iiit.ac.in}}
}

\maketitle

\begin{abstract}
To see what is not in the image is one of the broader missions of computer vision. Technology to inpaint images has made significant progress with the coming of deep learning. This paper proposes a method to tackle occlusion specific to human faces. Virtual presence is a promising direction in communication and recreation for the future. However, Virtual Reality (VR) headsets occlude a significant portion of the face, hindering the photo-realistic appearance of the face in the virtual world. State-of-the-art image inpainting methods for de-occluding the eye region does not give usable results. To this end, we propose a working solution that gives usable results to tackle this problem enabling the use of the real-time photo-realistic de-occluded face of the user in VR settings. 

This is an extended abstract of our CRV 2022 poster paper~\cite{ours}. 
\end{abstract}

\section{Introduction}
\label{sec:intro}

Face is a principle in any dialogue. Modern technology in the face-to-face conversation has gone from video to virtual presence over a distance. With the coming of Virtual Reality (VR), which promises to provide realism, most technology has been developed around creating a virtual \textit{avatar}. Nonetheless, using an avatar reduces the immersion and the sense of self-bringing in an uncanny valley. \cite{7223378} shows that increasing realism in the avatar does not improve the experience. The very construct of a VR headset restricts the use of the face, hindering the real-time broadcast of the user's face.

VR environment improves immersion and user experience~\cite{8263407}, but the occlusion of the face brings down this potential. Existing methods of reconstructing the face while using a VR device do not give usable results. The primary research challenge with the face completion/inpainting task comes from its ill-posed nature as a significant part of the face is occluded by HMD. Learning a single common face de-occlusion network with the capability to hallucinate diverse expressions in varying appearances and head poses across a large set of human faces is difficult to achieve. It is due to the broad space of facial geometry and appearance as well as the highly subjective way of articulating expressions/emotions across individuals~\cite{survey}. This motivates us to devise a person-specific model that gives usable results with high visual appeal and enables the use of the real-time photo-realistic inpainted face of the user in VR. 

\begin{figure}
\centering
    \includegraphics[width=\linewidth]{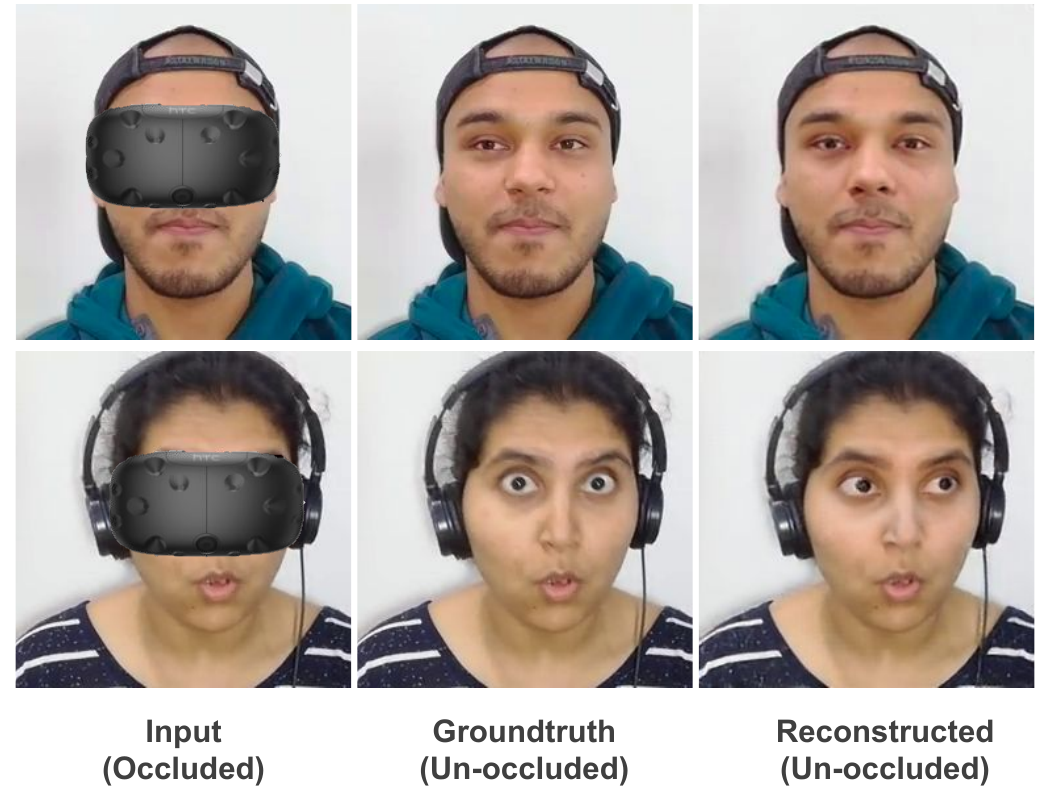}
    \caption{The proposed approach reconstruct unoccluded face image with high-quality even in presence of apparels.}
    \label{fig:teaser}
    \vspace{-1.5em}
\end{figure}



Traditional inpainting methods \cite{lafin,gc,ca} do not generalize to even slight non-frontal head poses and suffer from a loss of identity. Methods using an additional reference image to preserve identity also fail to scale beyond frontal head pose~\cite{genrgbd, faithful}. Elgharib~\etal shows that training a person-specific model is an effective way to model high-frequency details in generating frontal views from egocentric frames~\cite{egocentric}. 

As proposed, our method de-occludes face images to enable virtual presence without using 3D models. Previous methods that predict view conditioned texture and mesh geometry from HMD occluded input need calibrated multi-view data from the user, incurring high expenses for animation\cite{deep_app}. Figure~\ref{fig:teaser} shows high-quality de-occlusion results achieved by our method.

\section{Proposed Method}
We train a personalized model for face de-occlusion, particularly for application in VR teleconferencing, where the face is partially occluded due to HMD. We formulate the face de-occlusion problem as an image inpainting task. 

Given an input occluded face image $X_{occ}$, our network hallucinates the missing region with plausible and perceptually consistent facial details. We reconstruct the face image using the network output along with input image. The unoccluded image, $X_{rec}$ is compared against the ground truth unoccluded image, $X_{gt}$.



\subsection{Encoder-Decoder}

Our proposed architecture consists of an encoder-decoder module that comprises a stack of ResNet and inverted ResNet blocks. We use this framework to generate images using an adversarial loss. The encoder learns a 256-dimensional feature representation of the input image which is subsequently fed to the decoder network to reconstruct the target image.

\subsection{Attention Module}
\begin{figure}
\centering
    \includegraphics[width=\linewidth]{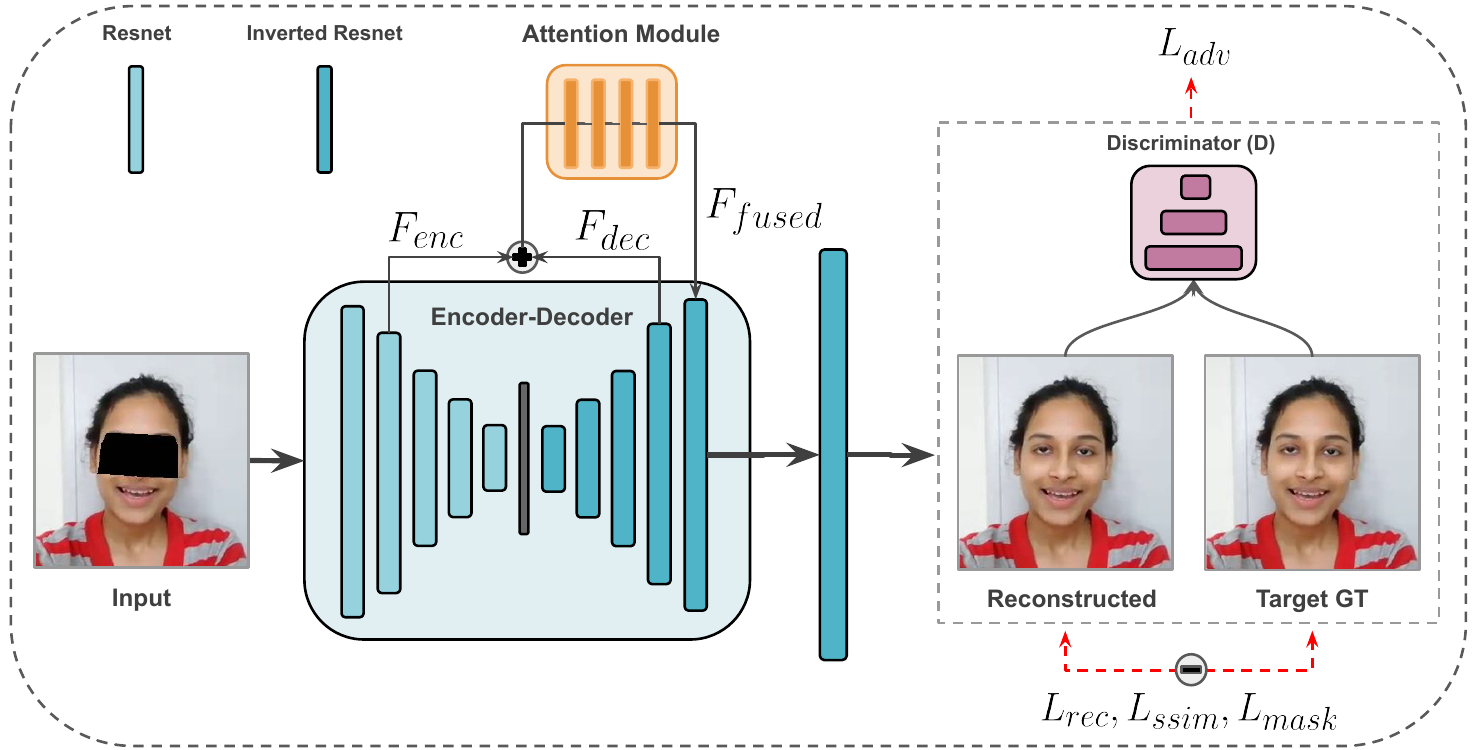}
    \caption{Detailed architecture diagram of our proposed network.}
    \label{fig:model}
    \vspace{-1.0em}
\end{figure}

We append our encoder-decoder module with an attention module to preserve the high-frequency appearance and background details. We perform spatial attention by taking the encoder output from the second layer, $F_{enc}$ of spatial dimension $64 \times 64$ and perform a channel-wise concatenation with the corresponding decoder layer output, $F_{dec}$ (i.e., with same spatial dimension). These are then fed to our attention module to generate attention maps of the same dimension as shown in Figure.~\ref{fig:model}. We then decouple these attention maps (i.e., $Attn_{enc}$ and $Attn_{dec}$) and use them for a weighted fusion of respective feature maps (i.e., $F_{enc}$ and $F_{dec}$) according to Equation~\ref{eq:attn}. The fused feature maps are fed downstream to convolution layers to reconstruct the de-occluded face image. 
\begin{equation}
F_{fused} = F_{enc}*Attn_{enc} + F_{dec}*Attn_{dec}
\label{eq:attn}
\end{equation}

\subsection{Loss Functions}
We employ a combination of four different loss functions as our training objective.
\begin{equation}
L_{rec} =  \lVert X_{gt}-X_{rec} \rVert_{1}
\label{eq:l1-loss}
\end{equation}
\begin{equation}
L_{adv} = log(D(X_{gt}))+log(1-D(X_{rec}))
\label{eq:adv-loss}
\end{equation}
\begin{equation}
L_{ssim} = SSIM(X_{rec},X_{gt}) 
\label{eq:ssim-loss}
\end{equation}

We also propose a novel \textit{mask-based loss} to further improve the quality of reconstruction in the HMD occluded area of the generated image and demonstrated generalization to unseen poses and appearances. Experiments show that our proposed method reports superior qualitative and quantitative results over state-of-the-art methods. 
\begin{equation}
L_{mask} = \lVert I_{mask}\odot I_{gt} - I_{mask}\odot I_{rec} \rVert_{1}
\label{eq:mask-loss}
\end{equation}
where, $I_{mask}$ refers to single channel binary mask image. $L_{rec}$, $L_{adv}$, $L_{ssim}$ and $L_{mask}$ refers to reconstruction loss, adversarial loss, SSIM based structural similarity loss and mask loss respectively.
\begin{equation}
\begin{aligned}
L_{final} = \lambda_{rec}*L_{rec}+\lambda_{adv}*L_{adv}+\lambda_{ssim}*L_{ssim}\\
+\lambda_{mask}*L_{mask}
\end{aligned}
\label{eq:floss}
\end{equation}

where, $\lambda_{rec}, \lambda_{adv}, \lambda_{ssim}$ and $\lambda_{mask}$ are the corresponding weight parameters for each loss term. 

\subsection{Implementation Details}

Considering the architecture diagram in Figure~\ref{fig:model}, we first pre-train our model without attention on publicly available face datasets such as VGGFace~\cite{vggface2} and AffectNet~\cite{affectnet}. The motivation for the use of such datasets is to leverage the inherent knowledge about the face structure.
\vspace{-1em}

\subsubsection{Two-stage Training}
\begin{figure}
\centering
    \includegraphics[width=\columnwidth]{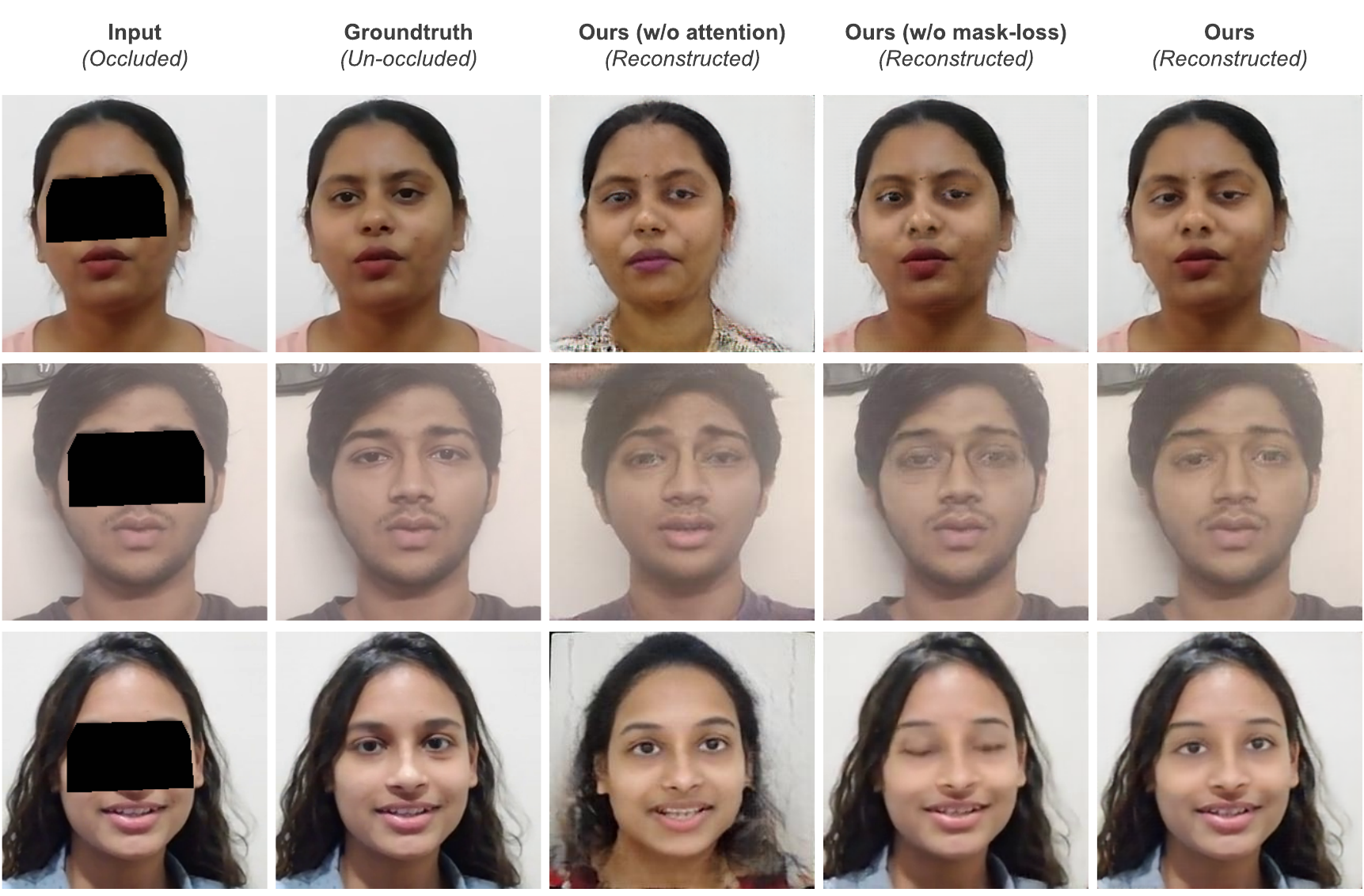}
    \caption{Visual results on ablation study. From left to right, third column shows results without attention and mask-loss, fourth column shows results with only attention and fifth column shows results with both attention and mask-loss.}
    \label{fig:inpaint-comp}
    \vspace{-1em}
\end{figure}

We further fine-tune our model in two stages in a person-specific setting. We train the first stage on unoccluded face images of the person to help the model focus on learning the user-specific facial structure, including eyes, nose, and mouth. In the second stage, we also train the attention module and fine-tune the entire model end-to-end on occluded face images of the same user. Figure~\ref{fig:inpaint-comp} justifies the importance of our attention module in capturing the high-frequency details, including appearance and background details. 

\subsubsection{Dataset}
To the best of our knowledge, there are no open-source datasets about faces with VR headsets. Hence, we captured various human subjects (around 20) in different appearances at a 1280 $\times$ 720 pixels resolution at 30 FPS from a mobile phone camera. We train our face de-occlusion network on  4-5 video sequences for each user, each length of around 1-2 min. Video frames are cropped and scaled to 256 $\times$ 256. We use mutually exclusive sets of these sequences from the same subject in different appearances to train and evaluate our user-specific model. Since we need ground-truth data during training, we overlay a synthetic binary mask over the eye region of unoccluded face images. However, we directly place it over the region occluded by the VR headset at test time. We use facial landmarks to guide the positioning of the mask. The structure of our mask imitates the look of a VR headset more closely than using rectangle boxes.

\section{Results and Discussion}
\begin{figure}
    \includegraphics[width=\columnwidth]{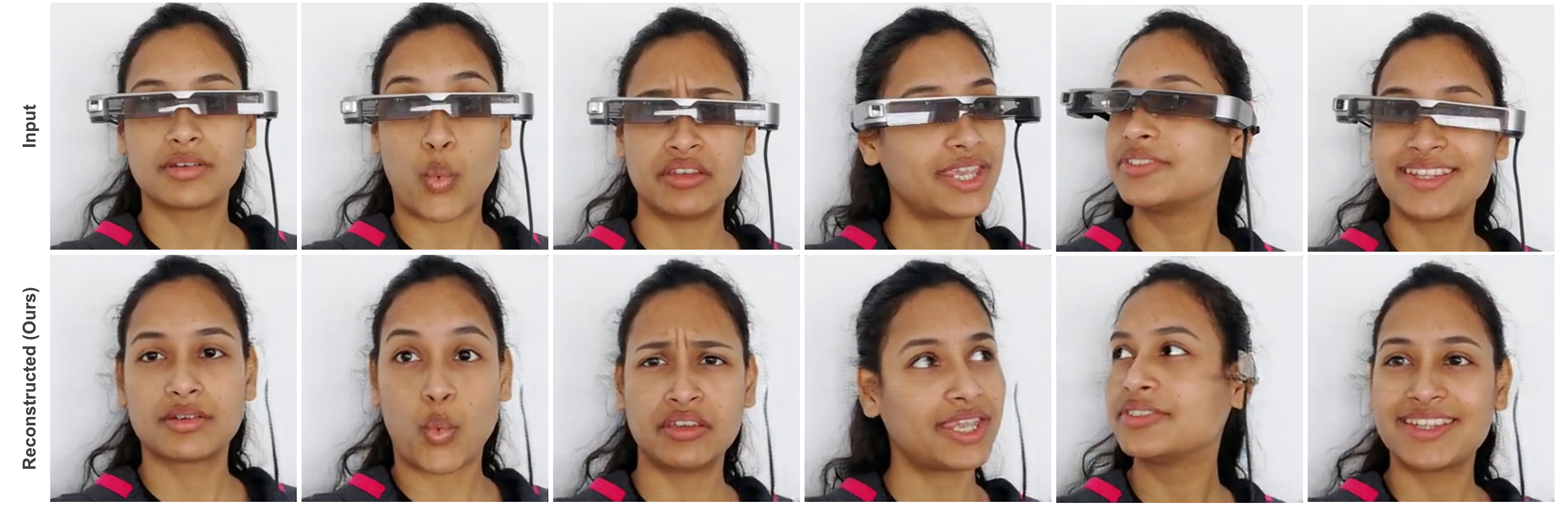}
    \caption{De-occlusion results using our method with large variations in head poses and expressions.}
    \label{fig:sg1}
\end{figure}

\begin{figure*}[!ht]
\centering
 \includegraphics[width=\textwidth]{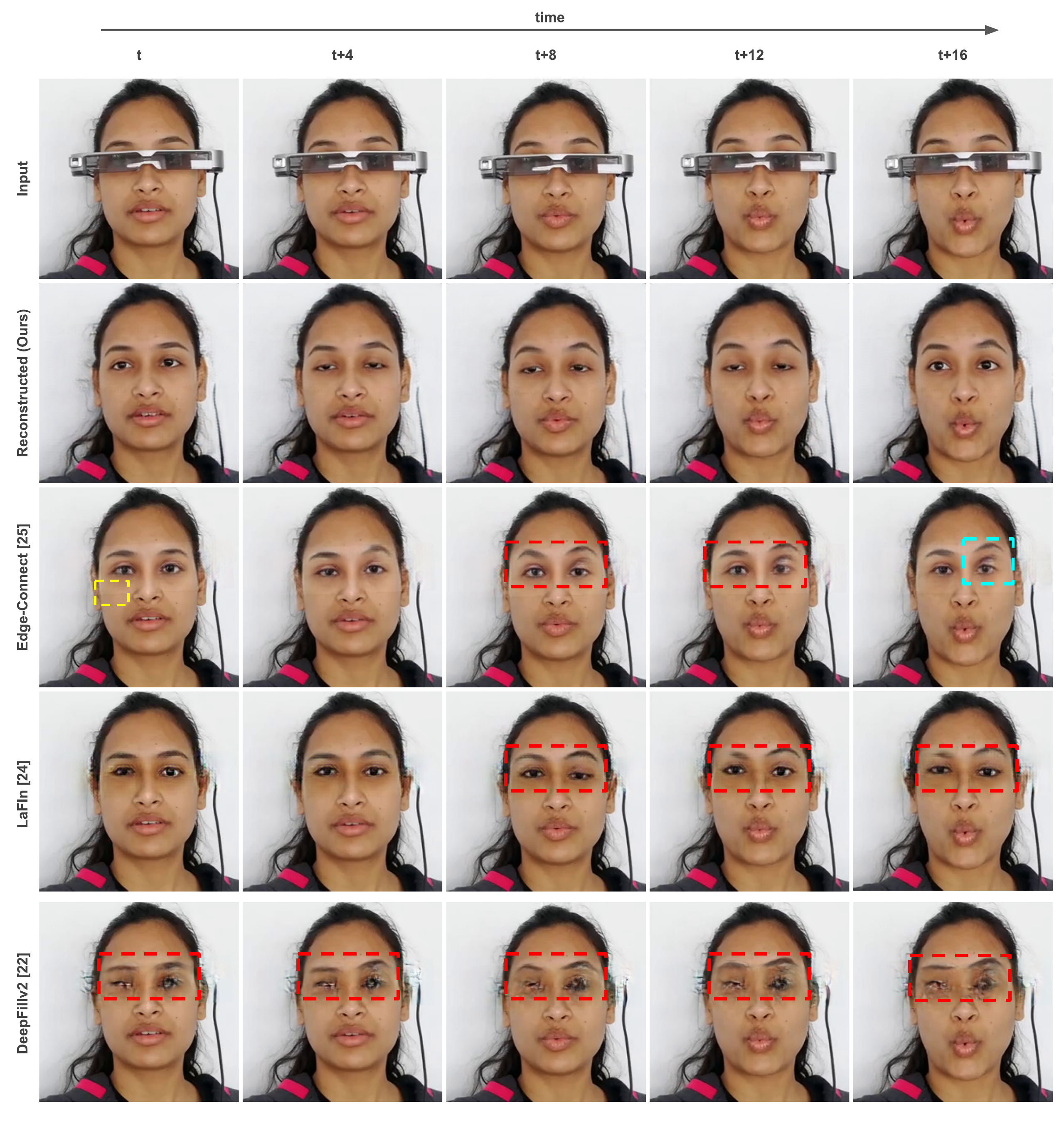}
    \caption{Qualitative comparison with other inpainting methods on real-world occlusion.}
    \label{fig:sg}
\end{figure*}

To justify the efficacy of our proposed approach, we conducted experiments on around 20 subjects and reported superior results in terms of all evaluation metrics. Qualitative and quantitative results in Figure~\ref{fig:sg} clearly show the importance of our person-specific training strategy over generalized inpainting methods such as DeepFillv2~\cite{ca}, LaFIn~\cite{lafin} and EdgeConnect~\cite{edgeconnect} that fail to generate realistic reconstructions across frames and suffers from visual artefacts. We also report de-occlusion results using our method in varying expressions and head poses in Figure~\ref{fig:sg1} to verify the generalizability of the proposed model.

\begin{table}[!ht] 
\centering
\caption{Quantitative comparison with other methods on face reconstruction.}
\label{table:compare}
\begin{tabu}{%
	l%
	*{7}{c}%
	*{2}{r}%
	}
\toprule
Method & SSIM$\uparrow$ &  PSNR$\uparrow$ & LPIPS$\downarrow$\\ 
\midrule
LaFIn~\cite{lafin} & 0.914& 23.693 & 0.0601  \\
EdgeConnect~\cite{edgeconnect} & 0.908 & 23.10 & 0.0689  \\
DeepFillv2~\cite{ca} & 0.845 & 19.693 & 0.117  \\
Ours (w/o attention) & 0.706 & 19.627 & 0.176 \\
Ours & 0.938 & 30.59 & 0.029 \\
\midrule
\end{tabu}%
\label{table: quant_res}
\vspace{-1em}
\end{table}

\begin{table}[!ht] 
\centering
\caption{Ablation study on different loss functions.}
\label{table:ablation}
\resizebox{1\columnwidth}{!}{ 
\begin{tabu}{%
	l%
	*{7}{c}%
	*{2}{r}%
	}
\toprule
Method & SSIM$\uparrow$ &  PSNR$\uparrow$ & LPIPS$\downarrow$\\ 
\midrule
Ours ($L_{rec}$)& 0.920 & 29.249 & 0.072 \\
Ours ($L_{rec}$+$L_{adv}$)& 0.822 & 27.638 & 0.141  \\
Ours ($L_{rec}$+$L_{adv}$+$L_{ssim}$) & 0.916 & 28.973 & 0.045\\
Ours ($L_{rec}$+$L_{adv}$+$L_{ssim}$+$L_{mask}$) & 0.918 & 29.025 & 0.042\\
\midrule
\end{tabu}%
}
\label{table: ablation}
\end{table}

\section{Applications}
In this paper, we address an ill-posed problem of facial de-occlusion in VR teleconferencing. Solving this problem is crucial to enabling many real-world applications. We can easily integrate facial animation models such as FOMM \cite{fomm, oneshot} in our setup to animate a face image from a reference image by just using sparse landmarks. Moreover, we can use this output animated face for per-frame 3D face reconstruction tasks ~\cite{df2net} and feed it to other VR teleconferencing users wearing a VR headset. Hence, our method allows VR and non-VR users to share a similar experience in a single hybrid teleconferencing application. For a demo example, refer to our supplementary video.

\section{Conclusion}
We aim to learn a personalized model for face de-occlusion in VR settings and formulate this problem as an inpainting task. Our proposed attention enabled encoder-decoder network takes an HMD occluded face as input and completes missing facial features, particularly the eye region. The experiments show that our method works reasonably well with the same person wearing different clothes, facial appearances, poses, and expressions. We believe that our approach can prove valuable for metaverse applications.

\clearpage
\clearpage
{\small
\bibliographystyle{ieee_fullname}
\bibliography{egbib}
}

\end{document}